\documentclass[runningheads]{llncs}
\usepackage{times}
\usepackage{epsfig}
\usepackage{graphicx}
\usepackage{amsmath}
\usepackage{amssymb}
\usepackage{booktabs} % for professional tables
\usepackage{soul}
\usepackage{xcolor}
\usepackage{multirow}
\usepackage{caption}
\usepackage{subcaption}
\usepackage{algorithm}
\usepackage{algorithmicx}
\usepackage{algpseudocode}

\usepackage[pagebackref=false,breaklinks=true,letterpaper=true,colorlinks,bookmarks=false]{hyperref}

\usepackage[T1]{fontenc}

\begin{document}
\title{The Importance of Image Interpretation: \\Patterns of Semantic Misclassification in Real-World Adversarial Images}
\titlerunning{Patterns of Semantic Misclassification in Real-World Adversarial Images}
\author{Zhengyu Zhao$^1$, Nga Dang$^2$, Martha Larson$^{1,2}$}
\authorrunning{Zhao et al.}
\institute{$^1$Institute for Computing and Information Sciences, Radboud University, Netherlands \\
$^2$Center for Language Studies, Radboud University, Netherlands
\email{z.zhao@cs.ru.nl,nga.dangthanhnga@student.ru.nl,m.larson@cs.ru.nl}\\
}
\maketitle              % typeset the header of the contribution
\begin{abstract}
Adversarial images are created with the intention of causing an image classifier to produce a misclassification.
In this paper, we propose that adversarial images should be evaluated based on semantic mismatch, rather than label mismatch, as used in current work.
In other words, we propose that an image of a ``mug'' would be considered adversarial if classified as ``turnip'', but not as ``cup'', as current systems would assume.
Our novel idea of taking semantic misclassification into account in the evaluation of adversarial images offers two benefits.
First, it is a more realistic conceptualization of what makes an image adversarial, which is important in order to fully understand the implications of adversarial images for security and privacy.
Second, it makes it possible to evaluate the transferability of adversarial images to a real-world classifier, without requiring the classifier's label set to have been available during the creation of the images.
The paper carries out an evaluation of a transfer attack on a real-world image classifier that is made possible by our semantic misclassification approach.
The attack reveals patterns in the semantics of adversarial misclassifications that could not be investigated using conventional label mismatch.
\end{abstract}
\section{Introduction}
\label{sec:intro}
An adversarial image is an image created to fool a classifier.
Researchers have shown that adversarial perturbations that are optimized with respect to one (known) model, the~\textit{source model}, can transfer their ability to cause a misclassification to another (unknown) model, the~\textit{target model}~\cite{papernot2016limitations,liu2017delving}.
Transfer attacks have been extensively studied in the non-targeted setting~\cite{dong2018boosting,dong2019evading,xie2019improving} as well as the more challenging, targeted setting~\cite{li2020towards,zhao2021success}.
Adversarial images that are \emph{transferable} in this respect have serious implications for the real world since they can be created by an external adversary, i.e., someone without insider knowledge of the target model.
On one hand, adversarial images, and especially transfer attacks, pose a danger to the security of systems, but, on the other hand, adversarial techniques can be used to protect privacy-sensitive content in images. %(add cite), 

\begin{figure}[!t]
\centering
\begin{subfigure}[b]{0.48\columnwidth}
\includegraphics[width=\columnwidth]{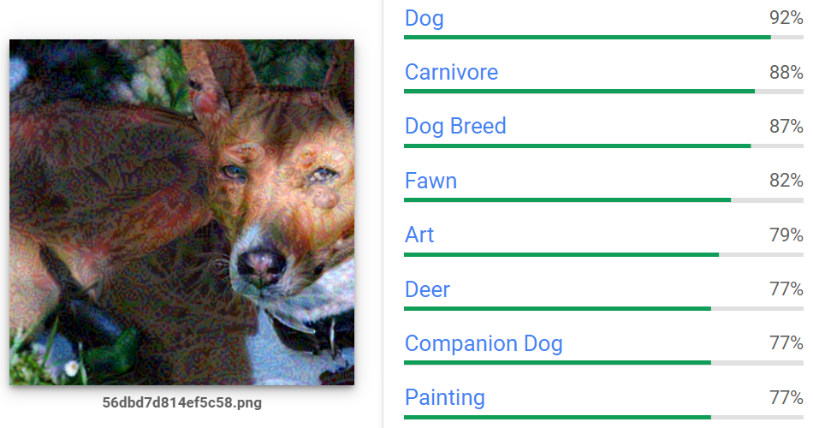}
\caption{Label misclassification}
\end{subfigure}
\begin{subfigure}[b]{0.48\columnwidth}
\includegraphics[width=\columnwidth]{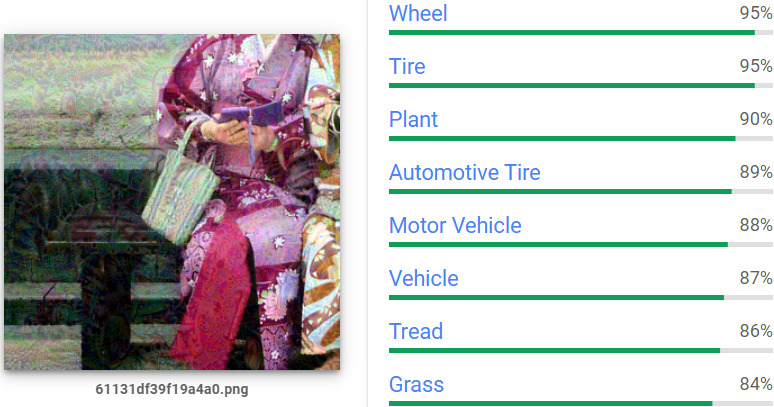}
\caption{Semantic misclassification}
\end{subfigure}
\caption{Adversarial images that fool the Google Cloud Vision (label detection). We propose that images that cause semantic misclassification (b) should be considered adversarial, but not images that cause label mismatch (a), as is currently the convention.}
\label{fig:exam}
%\vspace{-0.3cm}
\end{figure}

This paper is motivated by our observation that the current evaluation practices for adversarial images are not suited for real-world scenarios. 
Fig.~\ref{fig:exam} illustrates the issue.
The figure shows two adversarial images and the labels they have been assigned by the Google Cloud Vision API\footnote{\url{https://cloud.google.com/vision}}. 
The ground truth of the left image is ``basenji'', which is a breed of dog, and the ground truth of the right image is ``kimono'', which is the national dress of Japan.
Under conventional assumptions, these are both successful adversarial images since in neither case does the classifier predict the exact ground truth of the image.
From the point of view of human interpretation, however, these two cases are different.
The predicted labels for the left image are still clearly related to dogs; however, the predicted labels for the right image have nothing to do with the semantic content of the image.
In this paper, we propose the novel idea that the evaluation of adversarial images should take the semantic mismatch between the ground truth and the predicted labels into account.
In other words, it is not enough that the ground truth ``basenji'' is a label mismatch with the predicted label ``dog''. 
Rather, an adversarial image should only be considered successful misclassification if the ground truth and the predicted label are semantically unrelated.

The idea of evaluating adversarial images based on semantic misclassification has two benefits that are important in evaluating real-world systems.
First, it offers a more realistic conceptualization of what makes an image adversarial, with a better correspondence to human interpretation, as just discussed.
Previous work~\cite{kurakin2017adversarial,li2019scene,zhao2021success} has considered whether the mispredicted label was technically a difficult label, but has not considered human interpretations of which misclassifications are important.
Second, it makes it possible to evaluate the effectiveness of adversarial images in causing a real-world classifier to misclassify.
Existing work on adversarial images, e.g.,~\cite{dong2018boosting,inkawhich2020perturbing,li2020towards,xie2019improving,zhao2020towards,zhao2021success}, assumes access to the full label set used to train the target classifier.
In real-world evaluation, we may not have complete knowledge of this set.
For this reason, it is not possible to be certain that the ground truth label of the adversarial image can be possibly predicted by the classifier. 
Further, for so-called targeted adversarial images, where the classifier must output a pre-specified incorrect target label for the adversarial image to be considered successful, it is unknown if the target labels are in the label set of the classifier.
In these cases, conventional label mismatch evaluation is not meaningful and our semantic misclassification approach, which is based on human interpretation, is necessary.

To support our idea of evaluating adversarial examples with respect to semantic misclassification, in this paper we carry out a targeted transfer attack on a real-world classifier.
The results reveal interesting patterns of semantic misclassification that could not have been discovered with the current, conventional label mismatch evaluation approach.

\section{Experiments on Transfer Adversarial attacks}
\label{sec:back}
We experiment on two widely-used services of the Google Cloud Vision API: object detection and label detection.
Our experiments use the NIPS 2017 ImageNet-Compatible Dataset\footnote{Publicly available at~\url{https://github.com/cleverhans-lab/cleverhans/tree/master/cleverhans_v3.1.0/examples/nips17_adversarial_competition/dataset}.}, which consists of 1000 images with the size of $299\times299$.
Each image is associated with one of the 1000 ImageNet ground truth labels and one randomly assigned target label (used for the targeted attack).
Due to the time overhead for human judgments, we selected the first 400 images from the dataset for our experiments.
These 400 original images involve 266 unique ground truth labels and 310 unique target labels.

We experiment with iterative attacks that generate targeted adversarial images using pre-trained ImageNet classifiers as source models without additional data and model training.
We compare three algorithms for creating adversarial images with different loss functions: \emph{CE} (the widely used cross-entropy loss~\cite{kurakin2017adversarial}), \emph{Po+Trip} (an effective loss for targeted transferability, based on Poincar\'{e} distance and Triplet loss~\cite{li2020towards}) and \emph{Logit} (a state-of-the-art loss for targeted transferability, by solely maximizing the target logit value~\cite{zhao2021success}).
For each of these three attacks, we combine three widely-used transfer techniques: \emph{MI-FGSM} (uses a momentum term to accumulate previous gradients for more accurate updating~\cite{dong2018boosting}), \emph{TI-FGSM} (applies random translation to augment images for preventing the attack optimization from overfitting to the source model~\cite{dong2019evading}), and \emph{DI-FGSM} (applies random resizing and padding for image augmentation but also varies the augmentation parameters over iterations~\cite{xie2019improving}).

Following common practices, the perturbations were restricted by $L_{\infty}$ norm with $\epsilon=16$, and the step size of the attack optimization was set as 2.
To make sure all attacks can converge, we applied 300 iterations, following~\cite{zhao2021success}.
In order to further boost transferability, an ensemble of four pre-trained ImageNet classifiers (ResNet-50, DenseNet-121, VGGNet-16, and Inception-v3) is used as the source model.

\section{Evaluation based on semantic mismatch}
\label{sec:eval}

Evaluation was carried out by a human judge who applied our semantic misclassification approach.
The judge inspected and compared the predictions (object detection and label detection) of the images with the image ground truth.
Specifically, a non-targeted adversarial image was judged as successful if none of the top-10 predictions were semantically related to the ground truth label.
A targeted adversarial image was judged as successful if one of the top-10 predictions was semantically related to the specified target label for that image.\footnote{For reproducibility, all our human judgments are publicly available at~\url{https://github.com/ZhengyuZhao/Targeted-Tansfer/tree/main/human_eval}.}
In our evaluation, semantically related was taken to mean that the predicted label belonged to the broader semantic category of the ground truth label.
For example, the right-hand image in Fig.~\ref{fig:exam} is not a successful adversarial image under our evaluation approach since the predicted label ``dog'' is a broader semantic category of the ground truth label ``basenji''.

Note that we do not have a label set for the Google Cloud Vision API.
Consequently, it is not possible to evaluate using the conventional label-mismatch technique, as mentioned above.
Also, it is not possible to easily pre-judge or pre-compute the relations between labels and the broader semantic category that they belong to. 
We will return to the definition of semantic relatedness in our outlook in Sec.~\ref{sec:out}.

\begin{table}[!t]
\newcommand{\tabincell}[2]{\begin{tabular}{@{}#1@{}}#2\end{tabular}}
 \caption{Success rates (\%) of the three transfer attacks (CE, Po+Trip, Logit) on Google Cloud Vision.
 Both the targeted and non-targeted success rates are reported.
}
% \vspace{-0.5cm}
% \renewcommand{\arraystretch}{1}
\begin{center}
% \resizebox{\columnwidth}{!}{
\begin{tabular}{c|c|c|ccc}
\toprule[1pt]
Services&Evaluation&Ori&CE&Po+Trip&Logit\\
\midrule[1pt]
\multirow{2}{*}{\tabincell{c}{Object\\localization}}
&non-targeted&31.50&53.00&51.75&\textbf{62.50}\\
&targeted&0&9.00&8.50&\textbf{19.25}\\
\midrule[1pt]
\multirow{2}{*}{\tabincell{c}{Label\\detection}}
&non-targeted&9.75&34.00&22.50&\textbf{35.00}\\
&targeted&0&4.50&2.25&\textbf{6.25}\\
\bottomrule[1pt]
\end{tabular}
% }
\end{center}
 \label{tab:google}
%   \vspace{-0.5cm}

\end{table}

We collected prediction results through the simple process of taking screenshots of the Google Cloud Vision interface.
In each session, an image in its 4 (1 original and 3 adversarial) versions were uploaded, and then 8 screenshots of prediction results (for both object detection and label detection) were captured.
In total, for all the 400 different sessions of images, 3200 screenshots were gathered.
After optimization, it took about 4 minutes for each session, which amounted to 27-hour manual work of the human judge.
For both services, the prediction list consists of semantic classes along with confidence scores.
Note that the confidence score here is not a probability (which would sum to one).
Specifically, at most 10 classes with high confidence scores ($\geq$ 50\%) are returned. 

\section{Results}

The detailed attack results are reported in Table~\ref{tab:google}.
We can observe that, as expected, it is much more difficult for an attack to achieve targeted success than non-targeted success.
The Logit attack achieved the best performance in all cases.
This is consistent with the finding in~\cite{zhao2021success}.
When comparing the two services, we can observe that it is much easier to mislead object detection than label detection.

\begin{table}[!tp]
\newcommand{\tabincell}[2]{\begin{tabular}{@{}#1@{}}#2\end{tabular}}
\caption{\textbf{Left:} Quantity of predictions per image on different image sets (Ori, CE, Po+Trip, Logit) for both object detection and label detection services. \textbf{Right:} Diversity of predictions on different image sets in terms of the total number of unique predictions.}
%   \vspace{-0.5cm}
\renewcommand{\arraystretch}{1}
\begin{center}
\resizebox{\columnwidth}{!}{
\begin{tabular}{c|c|ccc}
\toprule[1pt]
Services&Ori&CE&Po+Trip&Logit\\
\midrule[1pt]
Object detection&2.58&2.92&2.68&3.37\\
\midrule[1pt]
Label detection&9.98&9.85&9.83&9.86\\
\bottomrule[1pt]
\end{tabular}
\quad
\begin{tabular}{c|c|c|ccc}
\toprule[1pt]
Services&All&Ori&CE&Po+Trip&Logit\\
\midrule[1pt]
Object detection&192&161&82&98&83\\
\midrule[1pt]
Label detection&1333&1162&730&771&1008\\
\bottomrule[1pt]
\end{tabular}
}
\end{center}
\label{tab:class_num}
    % \vspace{-0.2cm}

\end{table}

Our evaluation approach allows us to make some interesting observations about the semantic patterns of misclassifications.
Previously, such an analysis of a real-world system had not been possible.

We first looked at the coarse-grained impact of different attacks on the predictions.
To this end, we calculated the number of the returned predictions per image for original vs. adversarial images.
As can be seen from Table~\ref{tab:class_num} (left), for object detection, adversarial images yielded generally more predictions than original images, with the Logit attack yielding the most. 
In contrast, all results for label detection are very similar and close to the maximum returned number, 10.
This finding implies that label detection is more robust to adversarial attacks than object detection from a coarse-grained perspective.
Table~\ref{tab:class_num} (right) reports the total number of unique predictions for original vs. adversarial images.
As can be seen, for both two services, the predictions became less diverse after attacking.
This result is somewhat unexpected since there were 310 unique target labels (for the adversarial images), which are more than the 266 unique ground truth labels (for the original images).
This new finding implies that adversarial attacks are leading the classifier towards predicting a limited number of classes, which led us to further looking into the actual distribution of all unique predictions.

\begin{figure}[!t]
\centering
\includegraphics[width=\columnwidth]{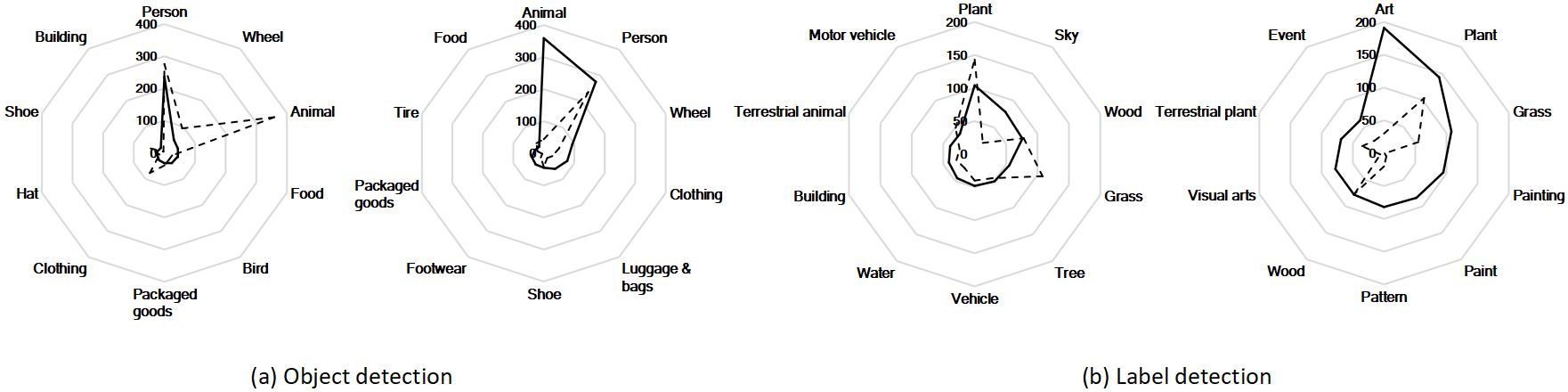}
\caption{Each figure shows the difference between the predictions on the original and the adversarial images (Logit).  For each of (a) and (b), the left-hand figure shows the Top-10 most frequent predictions for the original images (original predictions shown as solid line and adversarial as dashed line) and the right-hand figure shows the Top-10 most frequent for the adversarial images (adversarial predictions shown as solid line and original as dashed line).}
\label{fig:dis}
%\vspace{-0.3cm}
\end{figure}

Fig.~\ref{fig:dis} further visualizes the distributions of all unique predictions for original vs. adversarial images.
For object detection, the two image sets have different dominant classes.
People are frequently recognized in the original images, while animals become more frequent in the adversarial images.
For label detection, the added adversarial perturbations themselves seem to have led to labels related to art, such as ``art'' and ``painting'' or to vegetation, such as ``plant'' and ``grass'', being predicted.
This observation is also consistent with recent findings that adversarial images tend to be misclassified into some dominant classes that are visually similar to random perturbation patterns (e.g., ``brain coral'')~\cite{zhang2021data}.

\section{Conclusion and Outlook}
\label{sec:out}

In this paper, we have argued that conventional, label-mismatch, evaluation techniques do not support evaluation of the impact of adversarial images on real-world systems.
Instead, we have made the case that it is necessary to take human interpretation of what should be considered a semantic misclassification into account.
We have motivated our semantic-mismatch approach to evaluating adversarial images with examples and also with an analysis of the fooling ability of three different types of adversarial images on the Google Cloud Vision API.
Our approach has allowed us to identify patterns in misclassification.
In particular, transfer attacks can narrow the diversity of predicted classes and shift the semantics of the predictions in certain directions (e.g., towards animals or art).
We conjecture that these differences might be caused by specific properties of training data, or by design for specific purposes, e.g., outputting certain labels under conditions with low certainty. 
Investigations on how and why these differences exist would be promising for future work.

In this work, we consider a semantic mismatch to occur if the predicted label falls within a broader semantic category of the ground truth label.
However, it is important to note that our approach is not limited to this case.
Depending on the application scenario, it may be critical to ensure that the adversarial image pushes the classifier prediction even further from the semantics of the image.
For example, if someone was interested in sharing images taken in their bedroom and would like to ensure that large-scale image classifiers could not determine that the pictures were of the inside of their house, then adversarial versions of the images that cause a classifier to mistake ``bedroom'' for ``livingroom'' are not particularly useful and should not be considered successful.

\begin{figure}[!t]
\centering
\begin{subfigure}[b]{0.48\columnwidth}
\includegraphics[width=\columnwidth]{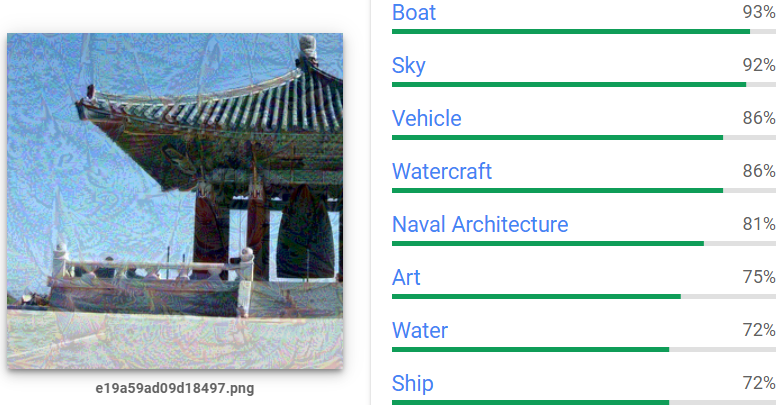}
%\caption{}
\end{subfigure}
\begin{subfigure}[b]{0.48\columnwidth}
\includegraphics[width=\columnwidth]{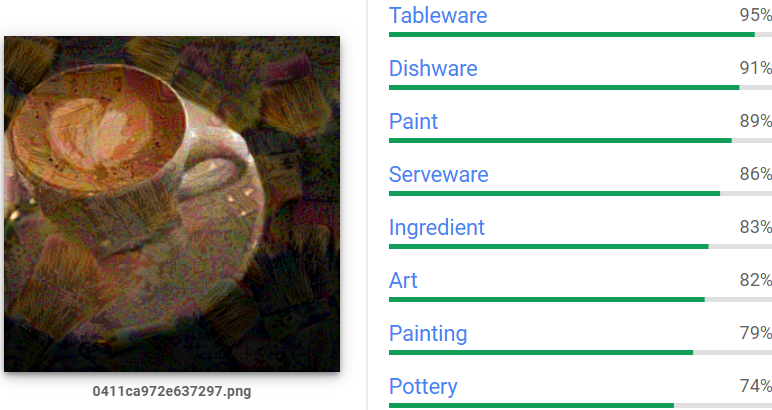}
%\caption{}
\end{subfigure}
\caption{Adversarial images that fool the label detection. These cases illustrate that human interpretation, with respect to a use scenario, is necessary to determine what should constitute a semantic mismatch, and thereby count as a successful adversarial image.}
\label{fig:more_exam}
%\vspace{-0.3cm}
\end{figure}

Fig.~\ref{fig:more_exam} shows two cases that illustrate that the human interpretation of semantic relatedness is dependent on what the most important content of an image is considered to be.
On the left, it is an image from our test set with the ground truth ``bell cote''. 
If this is indeed the most important content of the image, then it is a successful adversarial image since the predicted labels are not semantically related to bells or towers.
However, if sky is important then it is not.
On the right, the ground truth of the image is ``espresso''. 
Again, if this is indeed the most important content, then the image is a successful adversarial image because the predicted labels are not semantically related to coffee.
However, if the cup is important, then the image is not successful because ``tableware'', ``dishware'', and ``serviceware'' are predicted.
In short, it is evident that human interpretation is necessary in order to evaluate the success of adversarial images in a real-world use scenario.

To our knowledge, we are the first to provide human-related insights into the impact of transfer attacks on real-world computer vision systems.
It is important to note that further work is needed to understand how to implement human interpretation.
In addition to exploring other conceptualizations of semantic misclassification, future work should involve more annotators and conduct inter-annotator agreement analysis.

In sum, this paper lays the groundwork for a new line of research that gains a deeper insight into transfer attacks.
Moving forward, researchers should consider the human perspective on the change in the semantics of the prediction caused by adversarial images and should carry out analyses of the semantic patterns of predictions, in order to understand the impact of an attack at the system level in real-world scenarios.

% \clearpage
% \newpage

\bibliographystyle{splncs04}
\bibliography{ref}
% \end{thebibliography}
\end{document}